\let\accentvec\vec
\documentclass[runningheads, envcountsame, a4paper]{llncs}

\let\vec\accentvec
\pdfoutput=1
\newcommand{\doi}[1]{\textsc{doi}: \href{http://doi.org/#1}{\nolinkurl{#1}}}
\usepackage[pdftex]{graphicx}
\usepackage{epstopdf}

\usepackage{amssymb}
\usepackage{amsmath}

\usepackage[strings]{underscore}

\usepackage[usenames,dvipsnames]{xcolor} % useful as an alternative to comments
\usepackage[hyphens]{url}

\usepackage[caption=false]{subfig}

\begin{document}
\title{FaVoA: Face-Voice Association Favours Ambiguous Speaker Detection\thanks{The authors thank Leyuan Qu for the constructive feedback and suggestions, and acknowledge partial support from the German Research Foundation DFG under project CML (TRR 169).\protect\\The final authenticated version is available online at~\doi{10.1007/978-3-030-86362-3_36}.}}
%
%\titlerunning{FaVoA: Face-Voice Association Favours Ambiguous Speaker Detection}
%\author{Hugo Carneiro\inst{1}\orcidID{0000-0001-5094-5908} \and
%Cornelius Weber\inst{1}\orcidID{0000-0001-5163-938X} \and
%Stefan Wermter\inst{1}\orcidID{0000-0003-1343-4775}}
\author{Hugo Carneiro\inst{1} \and
Cornelius Weber\inst{1} \and
Stefan Wermter\inst{1}}
\authorrunning{H. Carneiro et al.}
\institute{Universität Hamburg, Department of Informatics, Knowledge Technology, 
Vogt-Koelln-Str. 30, 22527 Hamburg 
\email{\{hugo.cesar.castro.carneiro,cornelius.weber,stefan.wermter\}@uni-hamburg.de}\\
\url{http://www.informatik.uni-hamburg.de/WTM/}}
\maketitle              % typeset the header of the contribution
\begin{abstract}
The strong relation between face and voice can aid active speaker detection systems when faces are visible, even in difficult settings, when the face of a speaker is not clear or when there are several people in the same scene. By being capable of estimating the frontal facial representation of a person from his/her speech, it becomes easier to determine whether he/she is a potential candidate for being classified as an active speaker, even in challenging cases in which no mouth movement is detected from any person in that same scene. By incorporating a face-voice association neural network into an existing state-of-the-art active speaker detection model, we introduce FaVoA (\textbf{Fa}ce-\textbf{Vo}ice Association \textbf{A}mbiguous Speaker Detector), a neural network model that can correctly classify particularly ambiguous scenarios. FaVoA not only finds positive associations, but helps to rule out non-matching face-voice associations, where a face does not match a voice. Its use of a gated-bimodal-unit architecture for the fusion of those models offers a way to quantitatively determine how much each modality contributes to the classification.

\end{abstract}

\section{Introduction}
    \label{sec:introduction}
        
    \begin{figure}[hb!]
        \centering
        \includegraphics[width=0.95\textwidth]{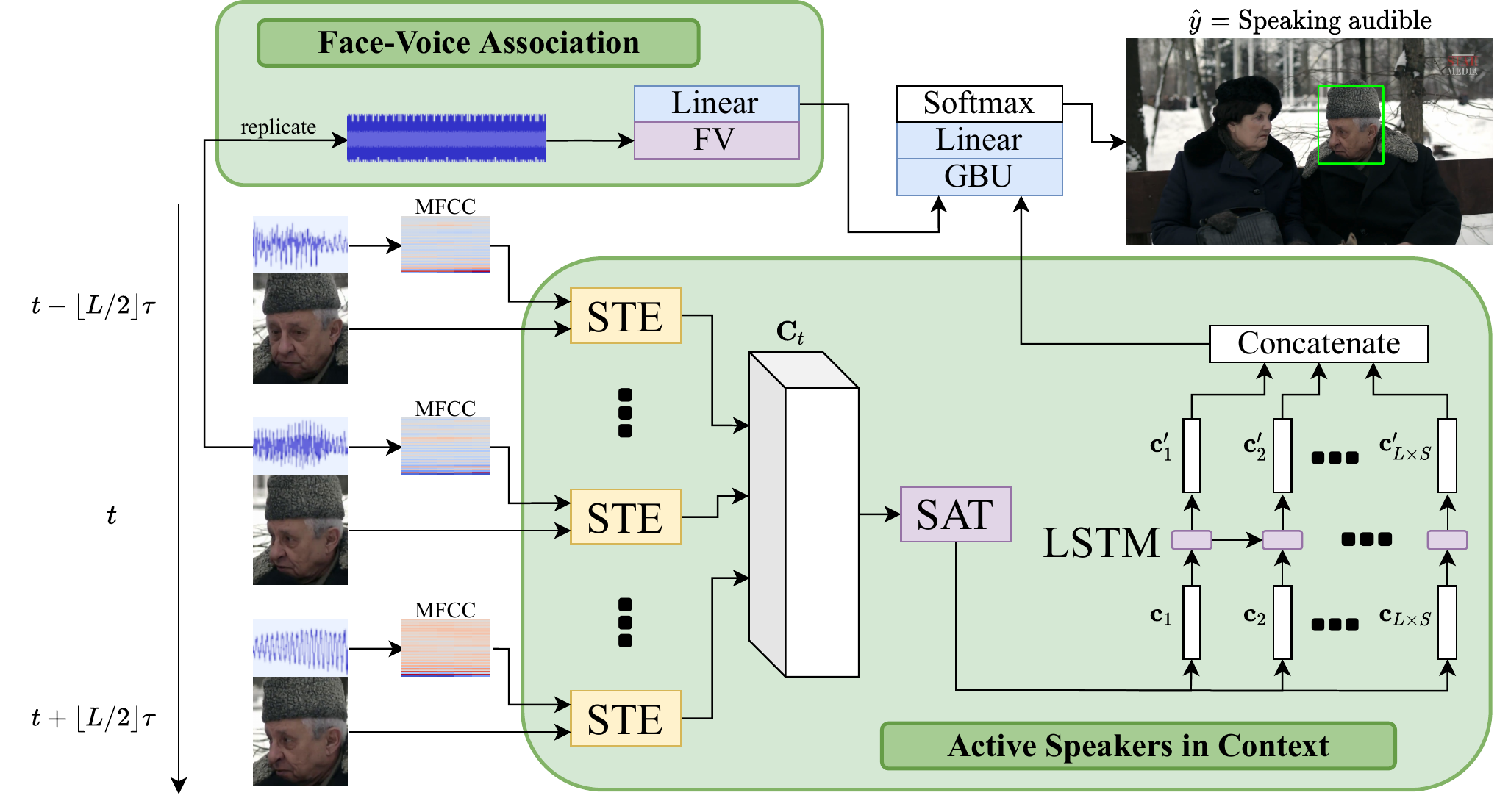}
        \caption{Active Speakers in Context (ASC) uses feature representations of face crops and audio provided by short-term encoders (STE). Through a pairwise analysis of feature representations of different speakers at distinct time steps made by a self-attention module (SAT) and the subsequent temporal refinement made by a long short-term memory (LSTM), ASC classifies an active speaker. By adding a face-voice association module (FV), FaVoA supports the classification of active speakers in challenging scenarios where the context does not provide enough information. The face-voice association module is combined with the output of ASC via a gated bimodal unit (GBU). Modules and layers in yellow are pretrained and fixed, those in violet are pretrained but are also updated during fine-tuning, and the ones in blue are trained from scratch.}
        \label{fig:model}
    \end{figure}
    
    The task of active speaker detection (ASD) consists of determining from which individuals in an audiovisual footage a given speaking activity originates. The combined use of auditory and visual modalities is fairly common in multimodal learning, including tasks like speech enhancement~\cite{hou2018}, speaker diarisation~\cite{chung2020}, speech reconstruction~\cite{qu2019}, and active speaker detection~\cite{alcazar2020b}. ASD is closely related to other audiovisual multimodal learning tasks, and a high-performing ASD model might help in paving the way for better models for those tasks to emerge. Related tasks include speech enhancement~\cite{hou2018} and speech separation~\cite{chung2020,qu2020}.
    
    Recent solutions to the problem of detecting speaking activity in the wild involve the use of 3D convolutions~\cite{chung2019,zhang2019b}, information from other individuals in the same scene~\cite{alcazar2020b} and the optical flow of facial movements~\cite{huang2020}. Although being very powerful, those models still face some difficulties depending on the resolution or the inclination of a person's face~\cite{alcazar2020b,huang2020,roth2020}. Most of them also struggle when working with medium- to long-term time spans~\cite{chung2019,huang2020,roth2020,zhang2019b}.
    
    In cases where faces are not clear enough, ASD must rely mainly on the auditory modality. However, in scenes where there are two people talking to each other and their faces are not clear enough -- due to a low resolution or to a high yaw inclination of their faces --, neither the visual nor the auditory modalities can provide enough information on their own. The existence of a module capable of retrieving a frontal face representation from the speaker's voice might provide information useful for speaker disambiguation in such challenging scenarios. Face-voice association applications show that it is actually possible to retrieve a frontal face representation from a speaker's speech signal~\cite{kim2018,oh2019}. The retrieved frontal face can be useful in cases in which the voice of the person speaking does not match the face of the person being classified for any of several reasons, e.g., difference in gender, ethnicity, age and so on. Additional information obtained via the crossmodal aspect of face-voice association, where one can relate one speech signal with a person's face, can help determining some clear cases that can be challenging for other models. For instance, if the mouth of the actual speaker in the scene is not seen for some reason, and no mouth movement is detected from any other participant in the scene. A non-speaking person whose face does not match the actual speaker's voice would be classified as not speaking. The actual speaker can also be properly classified if the face of no other scene participant matches the voice.
    
    The contributions of this paper include the creation of FaVoA (\textbf{Fa}ce-\textbf{Vo}ice Association \textbf{A}mbiguous Speaker Detector), a model (depicted in Figure~\ref{fig:model}) capable of detecting speaking activities in scenarios in which the context does not provide enough information, e.g., several people speaking simultaneously. We furthermore provide a quantitative evaluation on how much face-voice association actually contributes to the detection of speaking activity.
    
    The remainder of the paper is structured as follows. Section~\ref{sec:related} presents the approaches that have been proposed to tackle the active speaker detection task, as well as applications of face-voice association. Section~\ref{sec:model} introduces the model used in this research to address the task of active speaker detection. The model performance was assessed and compared with state-of-the-art architectures. The details on the experimental setup as well as its results are presented in Section~\ref{sec:experiments}. That section also offers a discussion on those results, as well as an analysis on how much importance face-voice association plays in ASD. Finally, Section~\ref{sec:conclusion} summarises the findings of this research and offers possibilities for future works.

\section{Related Works}
    \label{sec:related}
    
    \subsection{In-the-wild active speaker detection}
    \label{sec:asd}

        AVA-ActiveSpeaker~\cite{roth2020} was the first dataset built for in-the-wild active speaker detection. It was composed of videos in different resolutions with actors speaking in various distinct languages. Labels were provided for some speakers in selected frames of those videos depending on their speaking activity. The labels could be ``not speaking'', ``speaking audible'' and ``speaking not audible''. The dataset was built as part of a task at the 2019 ActivityNet Challenge. The task used mean average precision (mAP) as its evaluation metric and the audibly speaking activity as the positive class for that matter. Two competitors~\cite{chung2019,zhang2019b} achieved a higher mAP than the baseline provided by Roth et al.~\cite{roth2020}. Both models depended on a lip synchronisation preprocessing step, and could only achieve a high performance when working with short-term time spans and usually in scenarios in which there was only one person speaking~\cite{alcazar2020b,chung2019,zhang2019b}.
        
        To address the shortcoming of previous models, Alc\'{a}zar et al.~\cite{alcazar2020b} propose Active Speakers in Context (ASC), a model whose main intuition is to leverage active speaker context from long-term inter-speaker relations. It differs from previous approaches by using not only the information of the face of the target individual and of the audio input, but also that of the faces of other individuals detected at the same timestamp~\cite{alcazar2020b}. The addition of the information from the context in which a speaking activity happens grants ASC an mAP higher than that of Zhang et al.~\cite{zhang2019b}, but still lower than that of the ensemble models of Chung~\cite{chung2019}. Even though the context aids in some challenging scenarios, it may not prove useful in scenarios in which the mouth of the speaker is not seen due to low resolution or for the speaker not facing the camera, and when there are several people speaking simultaneously.
        
        Dense optical flow is also used for ASD, as a means to strengthen facial motion visual representation and this way avoid confusions that happen to audiovisual fusion-based models due to factors such as non-speaking facial motion, varied lighting and low-resolution footage~\cite{huang2020}. The inclusion of the dense optical flow grants the model a performance higher than the baseline model of Roth et al.~\cite{roth2020} in two distinct metrics~\cite{huang2020}, yet no mAP comparison is offered. No comparison with any other architecture is provided either. Similar to other models, the performance of that approach degrades when dealing with faces in low resolution or that are highly tilted.
    
    \subsection{Learning of face-voice association}
    \label{sec:facevoice}
        
        Learning of face-voice relations results from continuous and extensive exposure to audiovisual stimuli~\cite{gaver1993}. Psychology studies with infants indicate that the ability to make arbitrary face-voice associations emerge in humans between two and four months of age~\cite{bahrick2005}. In the area of active speaker detection, the advantage of matching visual and auditory representations was shown via the use of contrastive loss by some models~\cite{huang2020,zhang2019b}. Those implementations, however, do not explicitly make use of the advantages face-voice associations can provide.
        
        Applications of face-voice association in audiovisual crossmodal representation learning include the assembling of models capable of generating human faces from speech inputs~\cite{choi2020,oh2019}, as well as of models that can retrieve or match inputs from one modality given inputs of the other modality~\cite{kim2018,nagrani2018}. The performance of active speaker detection models degrades in cases where faces have a very small resolution or a large yaw angle~\cite{huang2020}. The ability to retrieve frontal facial embeddings from speech embeddings might provide additional information capable of helping with those challenging cases.
    
    \subsection{Gated bimodal unit}
    \label{sec:gmu}
        
        \begin{figure}[ht!]
            \centering
            \includegraphics[width=0.65\textwidth]{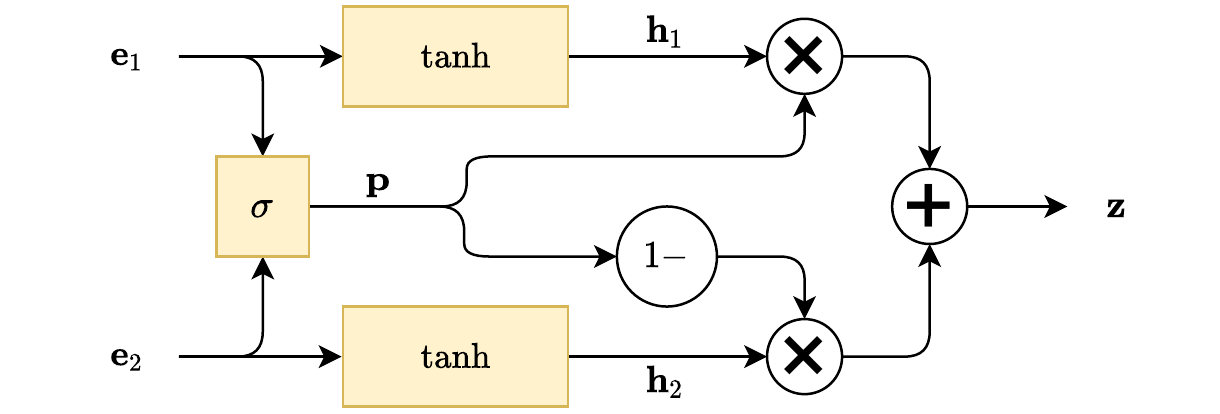}
            \caption{GBU inner structure}
            \label{fig:gbu}
        \end{figure}
        
        To determine if face-voice association presents an actual contribution to the task of ASD and in which cases it contributes the most, one should be able to evaluate its contribution quantitatively. Gated multimodal units (GMUs)~\cite{arevalo2017} are modality fusion mechanisms capable of providing quantitative values on the contribution of a given modality to the classification of a dataset entry. The gated bimodal unit (GBU) is a special case of the GMU oriented for the case where there are only two modalities to be fused. GMUs incorporate ideas from feature and decision fusion~\cite{arevalo2017}. The model architecture is based on the flow control of gated neural networks, e.g., gated recurrent units (GRUs)~\cite{cho2014}. Given embeddings $\mathbf{e}_1, \mathbf{e}_{2} \in \mathbb{R}^{d}$ from different modalities, the GBU outputs a fused embedding $\mathbf{z} \in \mathbb{R}^{d}$. As indicated in Figure~\ref{fig:gbu}, the GBU architecture is similar to the update gate of a GRU. In that sense, the GBU fused modality $\mathbf{z}$ is given by
        \begin{align}
            \mathbf{z} &= \mathbf{p} \odot \mathbf{h}_1 + (\mathbf{1} - \mathbf{p}) \odot \mathbf{h}_2, \label{eq:gbuz} \\
            \mathbf{p} &= \sigma \left( \mathbf{W}_p \left( \mathbf{e}_1 \mathbin\Vert \mathbf{e}_2 \right) + \mathbf{b}_p \right), \label{eq:gbup} \\
            \mathbf{h}_i &= \mathrm{tanh} \left( \mathbf{W}_i \mathbf{h}_i + \mathbf{b}_i \right),  \label{eq:gbuhi}
        \end{align}
        where $\odot$ denotes the Hadamard product, $\sigma$ the sigmoid function, $\mathbin\Vert$ vector concatenation and $\mathbf{1} \in \mathbb{R}^{d}$ an all-one vector. It is worth noticing from Equations~\ref{eq:gbuz} and~\ref{eq:gbup} that $\mathbf{p}$ can be interpreted as a vector of probabilities $p_1, p_2, \ldots, p_d$ that indicate the relevance of each modality in every element $z_i \in \mathbf{z}$. In other words, $z_i \in \mathbf{z}$ is composed of a linear combination of $h_{1, i} \in \mathbf{h}_1$ and $h_{2, i} \in \mathbf{h}_2$. The contribution of $h_{1, i}$ in $z_{i}$ is given by $p_{i}$ and that of $h_{2, i}$ is given by the complement of $p_{i}$, i.e., $1 - p_{i}$. Besides the case in which $p_{i} = 0.5$, one of the modalities will provide a major contribution to $z_{i}$ while the other will deliver a minor contribution.

\section{Model Architecture and Training Method}
    \label{sec:model}

    \subsection{Input data, Active Speakers in Context, and FaceVoice}
        \label{sec:input_data}
    
        FaVoA incorporates the context information of Active Speakers in Context (ASC)\hspace{0pt}~\cite{alcazar2020b} and the face-voice association provided by FaceVoice~\cite{kim2018}. And as such, the proposed model requires input data that can be fed to both models. Figure~\ref{fig:model} shows the architecture of the model, how it receives the input data and how it processes it. For the part imported from ASC, given a frame and a person in that frame, the model receives that person's face as a $144 \times 144$ image, as well as the audio input from that particular part of the video, which is converted to a MFCC spectrogram. Both inputs are sent to a short-term encoder, denoted as STE in Figure~\ref{fig:model}, which outputs a vector $\mathbf{u} \in \mathbb{R}^{1024}$. The STE is composed of two ResNet-18 CNNs~\cite{alcazar2020b}, one for each modality, which output vectors of 512 dimensions, which are then concatenated to produce $\mathbf{u}$~\cite{alcazar2020b}. The STE was pretrained with the weights provided by Alc\'{a}zar et al.~\cite{alcazar2020b} and kept fixed during training. From FaceVoice, only its voice subnetwork was used, which is denoted as FV in Figure~\ref{fig:model}. It requires 10 seconds of continuous speaking activity as input. However, it is not common for datasets built for active speaker detection to have the same person speaking for such a long time. To work around this restriction, the same audio input sent to STE was replicated until the repeated input had the length of 10 seconds. This approach was taken because the semantics of what is being said was irrelevant for this task and only the speaking activity was of interest. Given a 10-second audio input, FV then outputs a vector representation $\mathbf{a} \in \mathbb{R}^{128}$. FV was pretrained with the weights provided by Kim et al.~\cite{kim2018}, but unlike STE its weights were not kept fixed.
        
        In order to make use of the context in which a given speaking activity takes place, the vector representations $\mathbf{u}$ provided by the STE are combined and organised in a tensor $\mathbf{C}$. Tensor $\mathbf{C}$ is built in such a way that it may contain information from the time steps before and after the given speaking activity as well as from other speakers in the same scene. $\mathbf{C}$ has dimensions $L \times S \times 1024$, where $L$ is the number of frames used for the context and $S$ is the number of speakers. Those $L$ frames are defined according to a specific time step $t$, in which the speaking activity to be classified happens. The frames must be selected in a way that time step $t$ lies at the centre of the frame sequence. A sequence of $L$ frames should contain every frame from time step $t - \lfloor L / 2 \rfloor \tau$ to $t + \lfloor L / 2 \rfloor \tau$ with hops of $\tau$ units of time between each selected frame. It is worth noticing that the sequence of frames does not need to be contiguous. Given the frame of interest at time step $t$, a set of $S$ speakers in that frame is selected. If there are only $S^{\prime} < S$ speakers on the frame of interest, then information of some of those $S^{\prime}$ may be used more than once when working with that frame of interest. In a similar fashion, if some selected speaker appears only in a part of the frame sequence, its foremost activity is replicated all the way until the first frame of the sequence, and analogously its last activity is also replicated all the way until the last frame of the sequence. A more detailed explanation on the selection of frames and speakers can be found in the ASC original paper~\cite{alcazar2020b}. Tensor $\mathbf{C}$ is then subjected to a self-attention unit (SAT in Figure~\ref{fig:model}) and a single-layer LSTM for the sake of context refinement. The LSTM produces outputs $\mathbf{c^{\prime}_{i}} \in \mathbb{R}^{128}, 1 \leq i \leq L \times S$, which are concatenated into a vector representation $\mathbf{s} \in \mathbb{R}^{L \times S \times 128}$. SAT and LSTM were pretrained with the weights provided by Alc\'{a}zar et al.~\cite{alcazar2020b} and were subjected to updates during training.
    
    \subsection{Fusing speaking context and face-voice association}
        \label{sec:fusion_architecture}
        
        By combining the embedding $\mathbf{a}$, provided by FaceVoice, with $\mathbf{s}$, provided by ASC, it is expected that the benefits of face-voice association might aid the active speaker detection model even in cases in which the context is not enough, e.g., when there are several people speaking simultaneously, or when the faces of the speakers are either in low resolution or very tilted. The fusion of those embeddings is made by a GBU unit, but since it requires both modality embeddings to have the same dimension, embedding $\mathbf{a}$ is presented to a ReLU and a linear layer, which outputs a vector representation $\mathbf{a}^{\prime} \in \mathbb{R}^{L \times S \times 128}$. Both $\mathbf{a}^{\prime}$ and $\mathbf{s}$ are then fused by the GBU unit, which produces a fused vector representation $\mathbf{z} \in \mathbb{R}^{L \times S \times 128}$. 
        %FaVoA offers the prediction label $\hat{y}$ by applying a linear layer and a subsequent softmax operation to $\mathbf{z}$.
        The probability $q \! \left( \mathbf{x} \right)$ of a given input $\mathbf{x}$ being classified as ``speaking audible'' is obtained by projection from $\mathbf{z}$ with a linear layer and then the application of a softmax operation over the two classes.
        
        FaVoA was trained on a single NVIDIA GeForce RTX 2080 Ti GPU with 11 GB GDDR6 memory. A single cross-entropy loss $\mathcal{L}$ was used to train it using PyTorch. The loss is given by
        \begin{equation}
            \mathcal{L} = - y \log \! \left( q \! \left( \mathbf{x} \right) \right)  - \left( 1 - y \right) \log \! \left( 1 - q \! \left( \mathbf{x} \right) \right),
        \end{equation}
        where $y$ represents the expected label, which should be $1$ if there is audible speaking activity, and $0$ otherwise. The model weights were updated through a backpropagation algorithm, by trying to minimise the cumulative loss in every training mini-batch. Data was sent to the model via mini-batches of size 16. Similar to ASC, the model optimisation was done with the ADAM optimiser with an initial learning rate $\gamma = 3 \times 10^{-6}$ and learning rate decay $\eta = 0.1$ every $10$ epochs.

\section{Experiments}
    \label{sec:experiments}

    \subsection{Dataset}
        \label{sec:dataset}
    
        The AVA-ActiveSpeaker dataset is the first dataset intended for the task of active speaker detection that can be considered to be ``in the wild''. Prior to its publication, datasets crafted for this task were mainly composed of high resolution videos with the speakers facing the camera~\cite{roth2020}. AVA-ActiveSpeaker contains videos spoken in very distinct languages, with some of them with low resolution and with video and audio not well synchronised. Speakers may also appear in different video depths, which may cause facial information to be less clear for a learning system, and usually they are not looking at the camera.
        
        The AVA-ActiveSpeaker dataset contains 153 videos, split into 120 for training and 33 for validation. The training dataset is composed of 29,723 speaking/\hspace{0pt}non-speaking streams, ranging from 23 to 304 annotated entries, performed by a total of 10,156 distinct actors, some of them appearing in up to 2,165 dataset entries. The validation dataset has 8,015 streams of speaking/non-speaking activity that range from 14 to 305 dataset annotated entries. Those streams are captured from the performance of 2,515 distinct actors, with some of them having up to 2,143 entries of activity stored in the validation dataset. Table~\ref{tab:label_distribution} displays the label distribution among those datasets.
        
        \begin{table}[ht!]
            \caption{Label distribution of training and validation splits of the AVA-ActiveSpeaker dataset.}
            \label{tab:label_distribution}
            \centering
            \def\arraystretch{1.5}\tabcolsep=6pt
            \begin{tabular}{@{} l c c c @{}}\hline
                & \emph{Not Speaking} & \emph{Speaking Audible} & \emph{Speaking Not Audible} \\ \hline
                Training & $1,969,134$ & $682,404$ & $24,776$ \\
                Validation & $567,815$ & $192,748$ & $7,744$ \\ \hline
            \end{tabular}
        \end{table}
        
        \begin{table}[hb!]
            \caption{Comparison with state-of-the-art models on the validation subset.}
            \label{tab:comparison}
            \centering
            \def\arraystretch{1.5}\tabcolsep=6pt
            \begin{tabular}{@{} l c c c @{}} \hline
                & \textbf{mAP} $\uparrow$ & \textbf{AUC} $\uparrow$ & \textbf{Balanced accuracy} $\uparrow$ \\ \hline
                ASC~\cite{alcazar2020b} & $0.871$ & N/A & N/A \\ %$0.944$ & $0.831$ \\
                TC-LSTM Ensemble + Wiener~\cite{chung2019} & $0.878$ & N/A & N/A \\
                Multi-Task Learning~\cite{zhang2019b} & $0.840$ & N/A & N/A \\
                V+O+A VCE-CL~\cite{huang2020} & N/A & $0.932$ & $0.869$ \\
                AV-GRU-f2~\cite{roth2020} & $0.821$ & $0.910$ & $0.814$ \\ \hline
                \textbf{FaVoA} & $0.847$ & $0.928$ & $0.846$ \\ \hline
            \end{tabular}
        \end{table}
    
    \subsection{Experimental results}
        \label{sec:results}
        
        To evaluate FaVoA, its performance was compared with AV-GRU-f2, the baseline model provided by Roth et al.~\cite{roth2020}, ASC (Active Speaker in Context)~\cite{alcazar2020b}, Chung's TC-LSTM Ensemble + Wiener smoothing~\cite{chung2019} and Zhang et al.'s Multi-Task Learning model~\cite{zhang2019b}. Following the indications on the 2019 ActivityNet challenge, mAP is employed as the metric for this comparison. Table~\ref{tab:comparison} presents the achieved performance of state-of-the-art models and compares them with that of the model described in Section~\ref{sec:model}.
        
        Comparisons were also made with Huang and Koishida's F+O+A VCE-CL (Facial Image, Optical Flow and Audio Signal Visual-Coupled Embedding with Contrastive Loss)~\cite{huang2020}. It, however, does not offer performance values using the mAP metric. Because of this, a comparison is here provided using other metrics instead, namely the area under the ROC curve (AUC) and the balanced accuracy. Those metrics were also published for AV-GRU-f2~\cite{roth2020}. Performance results of those models, as well as FaVoA's, are also offered by Table~\ref{tab:comparison}. Table~\ref{tab:comparison} shows that not only FaVoA outperformed AV-GRU-f2 in every metric, but it also presents an mAP considerably higher than that of the multi-task learning approach~\cite{zhang2019b}, which was the runner-up in the 2019 ActivityNet challenge. Its AUC is also close to that obtained by V+O+A VCE-CL~\cite{huang2020}.
    
    \subsection{Contribution of face-voice association to active speaker detection}
    \label{sec:contribfacevoice}
    
        Ablation studies are performed to determine whether a given addition to a model makes an actual difference in its performance. However, they do not offer quantitative measures of how much that addition contributes to the classification. For multimodal classification, this is an important issue if one wants to better understand whether some modality contributes more than another to a given task. The use of GBU for crossmodal integration allows to determine if a given classification favours one modality or another. In the case of this study, the interest lies in determining if the classification is mostly due to context information (from ASC) or to face-voice association (from FaceVoice).
        
        In order to quantify the contribution of each modality, one can use the vector $\mathbf{p}$ produced by the GBU sigmoid unit (see Figure~\ref{fig:gbu} and Equation~\ref{eq:gbup}). For every entry of the dataset, a vector $\mathbf{p}$ can be extracted. This vector contains elements $p_i$, whose values range from $0$ to $1$. Each element $p_{i}$ represents a degree of contribution of modality input $\mathbf{e}_{1}$ (see Figure~\ref{fig:gbu}) to element $z_{i} \in \mathbf{z}$. In turn, the degree of contribution of modality input $\mathbf{e}_{2}$ to element $z_{i} \in \mathbf{z}$ is $1 - p_{i}$. By taking the fraction of elements of $\mathbf{p}$ whose value is greater than $0.5$, one can determine the fraction of elements of $\mathbf{z}$ that favours modality input $\mathbf{e}_{1}$ rather than $\mathbf{e}_{2}$. This way, one can get a quantitative measure of the contribution of modality input $\mathbf{e}_{1}$ to the classification and consequently, the contribution of $\mathbf{e}_{2}$ is simply one minus the contribution of $\mathbf{e}_{1}$. In our case modalities $\mathbf{e}_{1}$ and $\mathbf{e}_{2}$ correspond to the resulting vector representation of the FaceVoice module and the one of ASC.
        
\begin{figure}[hb!]
        	\centering
        	\subfloat[Number of entries per degree of contribution.\label{fig:contribution_graph}]{
        	    \raisebox{-.5\height}{
        		    \includegraphics[width=0.5125\textwidth]{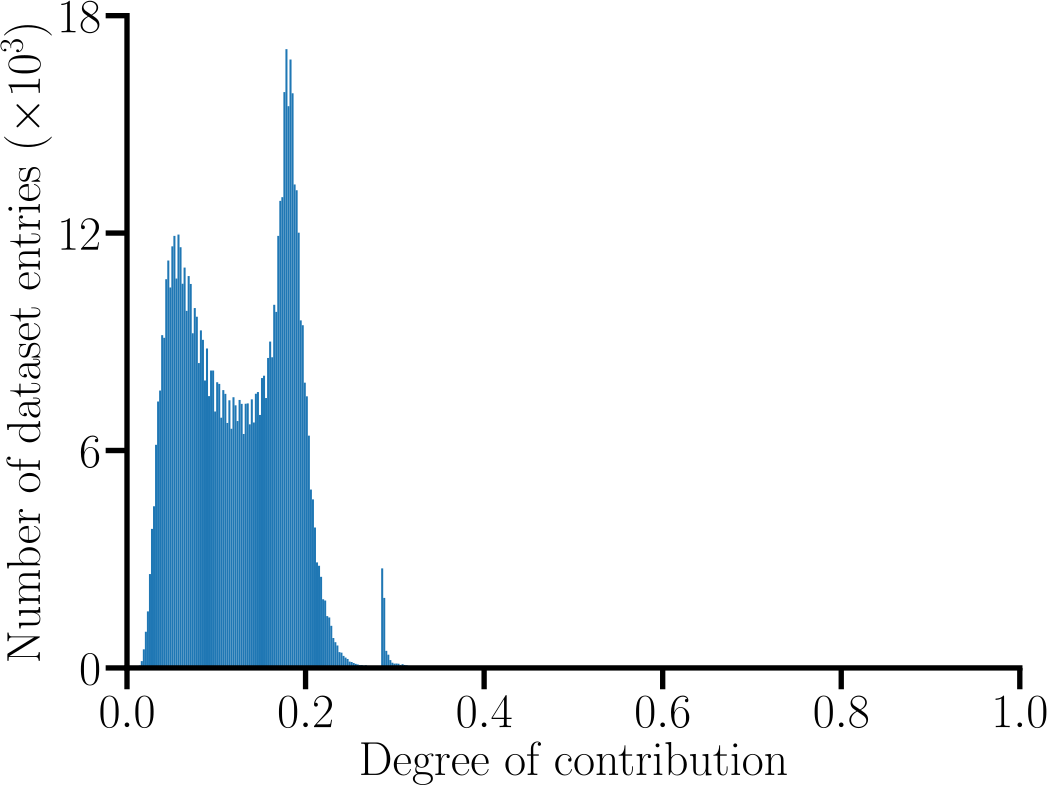}
        		    \rule[-1pt]{0pt}{50pt}
        		}
        	}\hfil
        	\subfloat[Face-voice association has a much higher degree of contribution for the man marked in green than for the other actors.\label{fig:gbucontribution_example}]{
        	    \raisebox{-.5\height}{
        		    \includegraphics[width=0.375\textwidth]{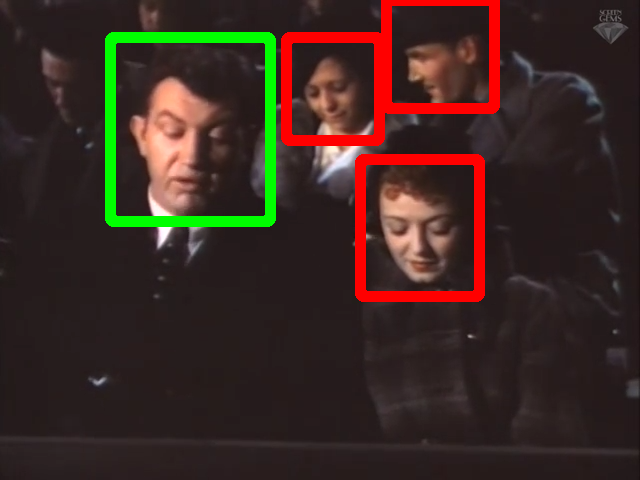}
        		    \rule[-18pt]{0pt}{50pt}
        		}
        	}
            \caption{Degree of contribution of face-voice association in entries of the validation set.}
            \label{fig:contribution}
        \end{figure}

        The graph of Figure~\ref{fig:contribution_graph} presents a histogram of the degree of contribution of face-voice association to the detection of speaking activity in entries of the validation set. The horizontal axis of the graph represents the degree of contribution of face-voice association, ranging from $0$ to $1$. The vertical axis represents the number of entries in the dataset for which the face-voice association had a particular degree of contribution. It can be noticed in the graph that context has a greater contribution to the classification than face-voice association in the entries of the validation set. Nevertheless, context is never favoured by all elements of the GBU output. Besides, face-voice association has a degree of contribution greater than $0.15$ for nearly $40\%$ of the entries, and for $303$ entries this degree of contribution can get higher than $0.3$.
        
        The contribution graph has three modes. The highest peak and its surrounding values correspond mostly to active speakers whose faces are clearly visible, or to silence. The region surrounding the leftmost peak corresponds to dataset records where there is some sound activity and the face of the active speaker is not entirely clear or the face being analysed is clearly not from the active speaker. The rightmost part of the graph, with degrees of contribution greater than $0.275$, corresponds to entries in which there are very loud sounds. Figure~\ref{fig:gbucontribution_example} depicts a scene in which the GBU assigns a reasonably higher degree of contribution of face-voice association for the man in the foreground ($0.198$) than for the other actors ($0.098$ for the woman in the foreground, and $0.129$ and $0.132$ for the actors in the background). This happens due to the presence of a male voice in the scene and the higher resolution of the face of the man in the foreground.
    
    \subsection{Comparison with Active Speakers in Context}
        
\begin{figure}[ht!]
        	\centering
        	\subfloat[Wrong gender.\label{fig:example_wrong_gender}]{
        		\includegraphics[width=0.3125\textwidth]{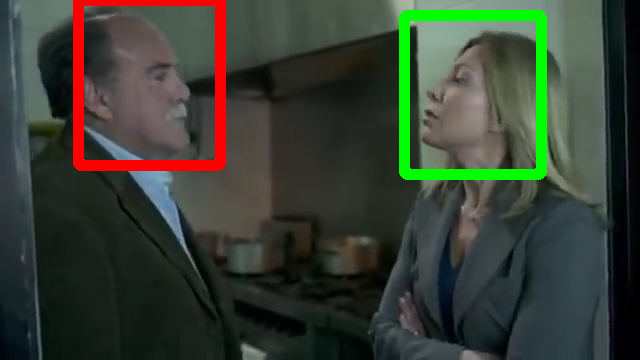}
        	}\hfil
        	\subfloat[Low resolution.\label{fig:example_low_resolution}]{
        		\includegraphics[width=0.3125\textwidth]{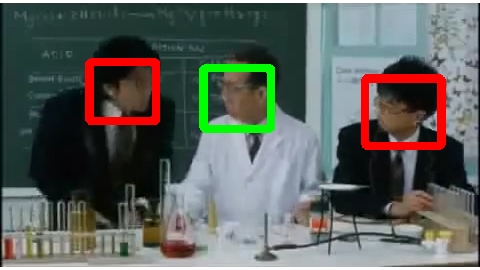}
        	}\hfil
        	\subfloat[Multiple speakers.\label{fig:example_multiple_speakers}]{
        		\includegraphics[width=0.3125\textwidth]{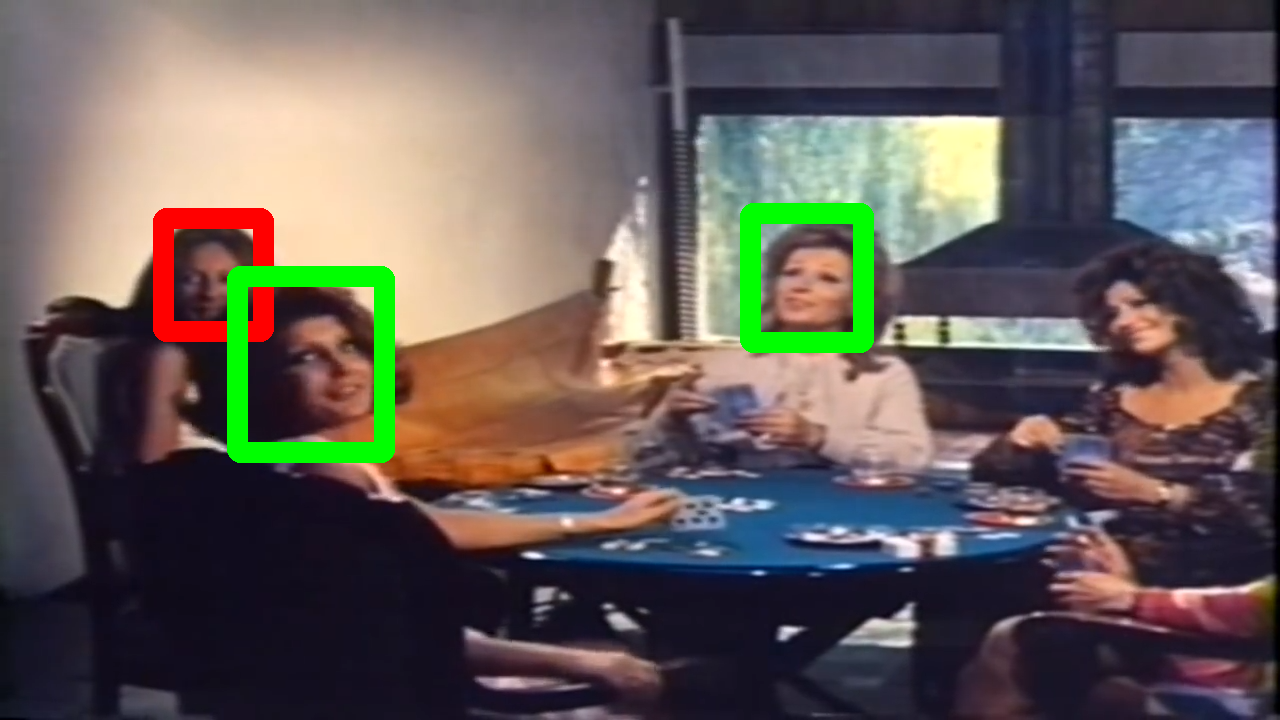}
        	}
            \caption{Examples of cases in which the context does not provide enough information and face-voice association is required for a correct active speaker detection. In the subfigures, people who are not speaking are marked with a red bounding box and those speaking with a green bounding box.}
            \label{fig:examples}
        \end{figure}

        The integration of FaceVoice into FaVoA offers the capability of classifying some instances of speech activity in which ASC failed. Figure~\ref{fig:examples} presents three cases in which the context information may be ambiguous and face-voice association proves useful. This may happen when actors are facing sideways and a facial feature may be mistaken for an open mouth. In Figure~\ref{fig:example_wrong_gender}, ASC wrongly classifies the facial hair for an open mouth, and classifies the man as speaking and the woman as not speaking. Face-voice association prevents this misclassification by recognising the female voice and associating it to the woman.
        
        ASC can also mistakenly classify speaking people as not speaking if the mouth of every person in the scene cannot be clearly seen due to low resolution (Figure~\ref{fig:example_low_resolution}) or if people are speaking simultaneously (Figure~\ref{fig:example_multiple_speakers}). ASC classifies every person in both figures as not speaking. Face-voice association can aid with correctly classifying the speaker of Figure~\ref{fig:example_low_resolution} due to the age difference. Regarding the scene depicted in Figure~\ref{fig:example_multiple_speakers}, ASC tends to classify a person as not speaking if someone in the same scene context seems to be already speaking. Thus ASC classifies both speaking women as not speaking, since the speaking activity of one of them triggers ASC to classify the other as not speaking and vice versa. Given the presence of female voices, FaVoA presents a less hesitant behaviour in classifying both women whose faces are not partially hidden as speaking.
        
        FaVoA presents some difficulties in comparison to ASC in scenes where the person is not speaking, but his/her voice can be heard narrating something. It also makes some mistakes in case there is some chanting and the voice of the person who is chanting somehow resembles that of the person being classified. Finally, ASC tends to more precisely classify some speaking activities (model outputs are mostly either close to $0$ or to $1$), whereas the outputs of FaVoA vary reasonably in the range between $0$ and $1$.

\section{Conclusion}
\label{sec:conclusion}
    
    This paper offers a study on the role of face-voice association in the task of active speaker detection. FaVoA provides a better classification in some challenging scenarios, such as low-resolution faces and several simultaneous speakers. Crossmodal learning models integrate the information from different modalities as a means to better tackle tasks in which one or more of those modalities do not provide enough useful information for some reason. By considering a person's characteristics by his/her voice, FaVoA makes use of the benefits of crossmodality in order to better determine the active speakers in a scene even in cases where the mouth of a speaker cannot be seen. The use of GBU for modality fusion allowed for determining quantitatively the contribution of face-voice association in ASD. By analysing that contribution, some cases of non-speaking activity can be immediately identified, which can help preventing the misclassification of some person as actively speaking. Cases in which there are several speakers can also be identified based on the degree of contribution of face-voice association. In future work, face-voice association may be used to support tackling other crossmodal tasks that involve conversational datasets in which speaker faces may not be clear. Additional directions for improvement in active speaker detection include the addition of other modalities, e.g., gaze and face keypoints.

\bibliographystyle{splncs04}
\bibliography{references}
%Author, F.: Article title. Journal \textbf{2}(5), 99--110 (2016)
%Author, F., Author, S.: Title of a proceedings paper. In: Editor,
%F., Editor, S. (eds.) CONFERENCE 2016, LNCS, vol. 9999, pp. 1--13.
%Springer, Heidelberg (2016). \doi{10.10007/1234567890}
%Author, F., Author, S., Author, T.: Book title. 2nd edn. Publisher,
%Author, A.-B.: Contribution title. In: 9th International Proceedings
%5on Proceedings, pp. 1--2. Publisher, Location (2010)
%LNCS Homepage, \url{http://www.springer.com/lncs}. Last accessed 4
%\end{thebibliography}
\end{document}